\documentclass[11pt]{article}

\usepackage{acl}
\usepackage{times}
\usepackage{latexsym}
\usepackage[T1]{fontenc}
\usepackage[utf8]{inputenc}
\usepackage{microtype}
\usepackage{xspace}
\usepackage{graphicx}
\usepackage{placeins}
\usepackage{amsmath,amssymb}
 \usepackage{booktabs}
\usepackage{xcolor}
\usepackage{array,colortbl}
\definecolor{swepreviewbasebg}{HTML}{EDF5FA}
\definecolor{swepreviewgain}{HTML}{39754B}
\definecolor{swepreviewdrop}{HTML}{B33D35}

\usepackage{listings}
\usepackage{hyperref}
\usepackage{url}
\usepackage{multirow}
\usepackage{float}
\usepackage{booktabs}
\usepackage{array}
\usepackage{tabularx}
\usepackage{multirow}
\usepackage{xcolor}
\usepackage{colortbl}
\usepackage{enumitem}

\providecolor{swepreviewbasebg}{HTML}{EDF5FA}
\definecolor{sweTabRed}{HTML}{C95550}
\definecolor{sweTabGreen}{HTML}{2E7D46}

\newlength{\sweTabValueWidth}
\newlength{\sweTabDeltaWidth}
\newlength{\sweTabHalfValueWidth}
\newlength{\sweTabCurrentWidth}
\newlength{\sweTabHalfCurrentWidth}

\providecommand{\sweTabDelta}[4]{}
\renewcommand{\sweTabDelta}[4]{%
  {\fontsize{4.8}{5.5}\selectfont
   \textcolor{#1}{\ensuremath{#2\mkern-4mu#3}\hspace{-0.08em}#4}}%
}

\providecommand{\sweTabCell}[2]{}
\renewcommand{\sweTabCell}[2]{%
  \settowidth{\sweTabCurrentWidth}{#1}%
  \setlength{\sweTabHalfCurrentWidth}{0.5\sweTabCurrentWidth}%
  \makebox[\linewidth][c]{%
    \makebox[0pt][c]{#1}%
    \if\relax\detokenize{#2}\relax\else
      \makebox[0pt][l]{\hspace{\sweTabHalfCurrentWidth}\hspace{0.16em}#2}%
    \fi}%
}

\newcommand{\sweTabSetDeltaWidth}[1]{%
  \settowidth{\sweTabDeltaWidth}{#1}%
}

\newcommand{\sweTabMainSetup}{%
  \small
  \settowidth{\sweTabValueWidth}{00.0}%
  \setlength{\sweTabHalfValueWidth}{0.5\sweTabValueWidth}%
  \setlength{\tabcolsep}{3pt}%
  \renewcommand{\arraystretch}{1.12}%
  \sweTabSetDeltaWidth{%
    \sweTabDelta{sweTabRed}{\downarrow}{-}{56.5}}%
}

\newcommand{\sweTabAblationSetup}{%
  \scriptsize
  \settowidth{\sweTabValueWidth}{00.0}%
  \setlength{\sweTabHalfValueWidth}{0.5\sweTabValueWidth}%
  \setlength{\tabcolsep}{1.8pt}%
  \renewcommand{\arraystretch}{1.05}%
  \sweTabSetDeltaWidth{%
    \sweTabDelta{sweTabRed}{\downarrow}{-}{29.8}}%
}

\newcommand{\sweTabOrderSetup}{%
  \scriptsize
  \settowidth{\sweTabValueWidth}{00.0}%
  \setlength{\sweTabHalfValueWidth}{0.5\sweTabValueWidth}%
  \setlength{\tabcolsep}{1.8pt}%
  \renewcommand{\arraystretch}{1.05}%
  \sweTabSetDeltaWidth{%
    \sweTabDelta{sweTabRed}{\uparrow}{+}{9.2}}%
}

\newcommand{\sweTabAlignSetup}{%
  \scriptsize
  \settowidth{\sweTabValueWidth}{00.0}%
  \setlength{\sweTabHalfValueWidth}{0.5\sweTabValueWidth}%
  \setlength{\tabcolsep}{1.8pt}%
  \renewcommand{\arraystretch}{1.06}%
  \sweTabSetDeltaWidth{%
    \sweTabDelta{sweTabGreen}{\uparrow}{+}{16.2}}%
}

\usepackage{amsmath,amsfonts,bm}

\def\eqref#1{(\ref{#1})}

\def\1{\bm{1}}

\DeclareMathAlphabet{\mathsfit}{\encodingdefault}{\sfdefault}{m}{sl}
\SetMathAlphabet{\mathsfit}{bold}{\encodingdefault}{\sfdefault}{bx}{n}

\definecolor{citeblue}{HTML}{5B9BD5}
\hypersetup{
  colorlinks=true,
  citecolor=citeblue,
  linkcolor=citeblue,
  urlcolor=citeblue
}

\graphicspath{{figures/}}

\title{SWE-Ladder: Learning to Localize, Reproduce, and Patch via\\
Progressive Privileged Self-Distillation}

\author{
  \textbf{Xudong Cai\textsuperscript{1} \quad Zhengxi Lu\textsuperscript{1} \quad
  Yuchen Yan\textsuperscript{1} \quad Jian Shao\textsuperscript{1}} \\
  \textbf{Jun Xiao\textsuperscript{1} \quad Yueting Zhuang\textsuperscript{1} \quad
  Yongliang Shen\textsuperscript{1}} \\
  \textsuperscript{1}Zhejiang University \\
  \texttt{\{ccdown, zhengxilu, syl\}@zju.edu.cn}
}

\newcommand{\syl}[1]{{#1}}

\newcommand{\method}{\textsc{SWE-Ladder}\xspace}

\lstdefinestyle{promptstyle}{
  basicstyle=\ttfamily\scriptsize,
  columns=fullflexible,
  keepspaces=true,
  breaklines=true,
  breakatwhitespace=false,
  showstringspaces=false,
  frame=none,
  aboveskip=0.4em,
  belowskip=0.2em
}

\begin{document}
\maketitle

\begin{abstract}
\syl{Resolving a repository-level issue typically requires a
software-engineering (SWE) agent to localize the faulty code,
reproduce the failure, and patch the code. Reinforcement learning
trains such agents from a single test result on the final patch,
while on-policy self-distillation (OPSD) adds token-level
supervision from a teacher with privileged evidence, 
exposing the full reference patch at every step. Neither provides
stage-specific guidance for localization or reproduction.
We propose \method, which turns the verified repair of each training
instance into stage-specific evidence: the changed files and lines
for localization, the fail-to-pass tests for reproduction, and the
diff for patching. A rule-based router assigns each step of an
end-to-end rollout to a stage based on its tool call, a frozen teacher
sees only the matching evidence, and the stages are distilled in
repair order while GRPO optimizes the full trajectory.
With Qwen3-4B-Instruct-2507 on SWE-bench Verified under the R2E-Gym
scaffold, \method improves pass@1 from $9.6\%$ to $16.4\%$, compared
with $10.4\%$ for GRPO and $12.4\%$ for OPSD with the full patch.
Under the OpenHands scaffold, it improves pass@1 from $13.0\%$ to
$17.6\%$. Reversing the stage order under R2E-Gym reduces pass@1 to
$11.4\%$.} Code is available at
\url{https://github.com/ZJU-REAL/SWE-Ladder}.
\end{abstract}

\section{Introduction}
\label{sec:intro}

\begin{figure}[t]
\centering
\includegraphics[
width=\columnwidth
]{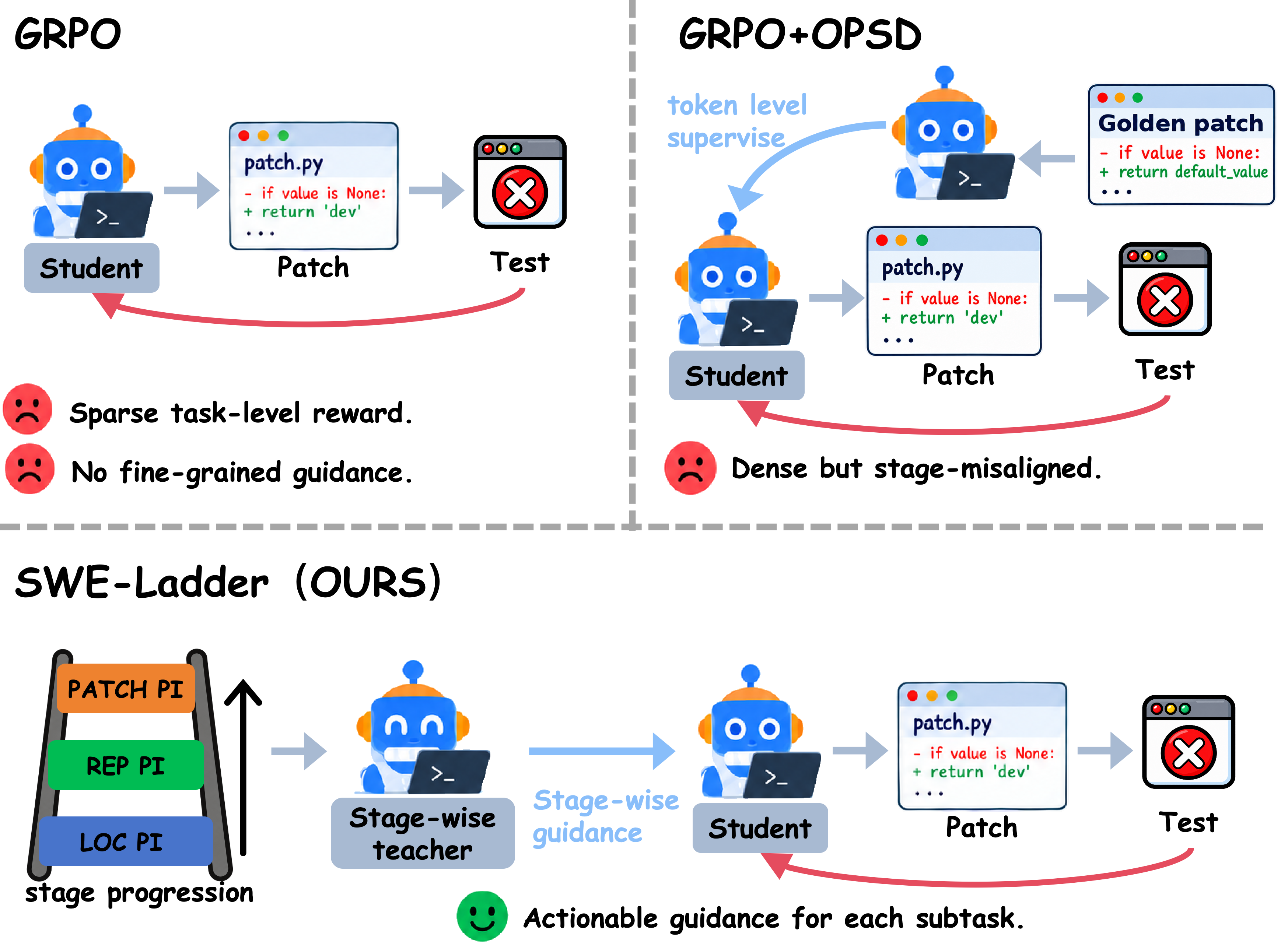}
\caption{\syl{\textbf{Supervision signals compared in this paper.} GRPO assigns one reward to the whole trajectory; GRPO+OPSD shows the teacher the full reference patch at every step; \method gives the teacher stage-specific evidence and introduces the stages one at a time.}}
\label{fig:privilege-motivation}
\end{figure}


\begin{figure}[t]
\centering
\includegraphics[
width=\columnwidth
]{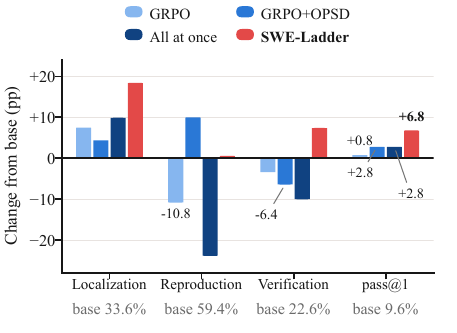}
\caption{\textbf{Stage-wise diagnostics on SWE-bench Verified.} \syl{Bars show changes from the base model at the final checkpoint; labels mark the changes discussed in Section~\ref{sec:intro}.} Base values appear under each metric. Definitions are in Appendix~\ref{app:stage-capability}.}
\label{fig:stage_capability}
\end{figure}

Reinforcement learning with verifiable rewards (RLVR) has become an effective approach to training software-engineering (SWE) agents~\citep{deepswe,longcontextswe}. The agent edits a repository over many tool calls, and the only training signal is whether its final patch passes the designated tests~\citep{swebench,swebenchverified}. Over dozens of turns, it is difficult to identify which decision caused a failure~\citep{agentless}. On-policy self-distillation (OPSD) offers denser supervision~\citep{opsd,sdpo}. The same model acts as a teacher that receives privileged evidence withheld from the student and scores every token of the student's trajectory~\citep{opd_tm,gkd}. In SWE, the available evidence is the verified repair of each training instance, and the natural choice is to show the teacher the full reference patch at every step \syl{(Figure~\ref{fig:privilege-motivation})}.

To see what these signals teach, \syl{we measure three intermediate behaviors} (Figure~\ref{fig:stage_capability}). They record whether the agent edits the files changed by the reference repair, reproduces the failure with a script, and re-runs the script successfully after its last edit. We find two patterns. First, the outcome reward does not consistently improve these behaviors: under GRPO, pass@1 barely moves (from 9.6\% to 10.4\%), while reproduction falls from 59.4\% to 48.6\%. Second, showing the teacher the full patch raises pass@1 but lowers post-edit verification from 22.6\% to 16.2\%.

The full patch conflates the evidence for three different skills. Localization needs to know where the fault is, reproduction needs to know how the failure shows up, and patching needs to know what to change. During a localization step, a teacher that sees the full patch receives not only localization cues but also patch-generation details, and may therefore favor editing over searching. A similar effect has been reported in single-turn reasoning, where a teacher conditioned on the full reference makes the student skip intermediate steps and verify less~\citep{aropd}. \syl{\emph{At each step, the teacher should therefore see only the part of the verified repair that the current stage needs.}}

The three skills support one another: localization identifies the code to investigate, reproduction makes the failure concrete, and the resulting test provides a basis for checking a patch. We therefore organize privileged supervision into a progressive training schedule: localization, then reproduction, then patch generation.

Following these observations, we propose \method, which trains a single end-to-end agent with progressive privileged self-distillation. Prior work obtains stage-level supervision by fixing a localize-then-repair pipeline at inference time~\citep{agentless,sorft,swe_swiss}, which limits how the agent can explore. \method leaves the agent's actions unconstrained and decomposes only the teacher's evidence. A deterministic router labels each step as localization (\textsc{LOC}), reproduction (\textsc{REP}), or patch generation (\textsc{PATCH}) from its tool call. The teacher then sees only the slice of the repair for that stage. Training proceeds through the three stages in order, distilling one stage at a time, while GRPO optimizes the full trajectory.

With Qwen3-4B-Instruct-2507 on SWE-bench Verified, the gain depends on how the teacher receives evidence. Without privileged evidence, GRPO improves pass@1 from 9.6\% to 10.4\%. With the full patch at every step, GRPO+OPSD reaches 12.4\%, and training all three stages at once also reaches 12.4\%. Introducing the stages in order, \method reaches \textbf{16.4\%} and is the only method that improves both localization and verification without lowering reproduction (Figure~\ref{fig:stage_capability}). \syl{Reversing the order lowers pass@1 to 11.4\%, and on the OpenHands scaffold \method improves pass@1 from 13.0\% to 17.6\%.}

Our contributions are threefold: \begin{itemize}[itemsep=1pt,topsep=2pt,parsep=0pt] \item We show with stage-wise diagnostics that outcome rewards do not consistently improve intermediate repair behavior and that \syl{a full-patch teacher} lowers post-edit verification.

\item We propose \method, which shows the teacher one stage's slice of the verified repair at a time, following the order of repair.

\item \method outperforms GRPO and full-patch GRPO+OPSD \syl{on two agent scaffolds}, and removing any stage\syl{, reversing the order,} or training all stages at once lowers pass@1. \end{itemize}

\section{Related Work}

\subsection{Post-Training for SWE Agents}
Recent datasets and environments expand the supply of executable
repair tasks, trajectories, and verifiers for SWE
training~\citep{swegym,r2egym,swesmith,swerebench,sweuniverse,bugpilot}.
SWE-RL applies GRPO with patch-similarity rewards, DeepSWE uses
verifier-based RL, and long-context SWE training combines rejection
fine-tuning with multi-turn RL~\citep{swerl,deepswe,longcontextswe}.
Kimi-Dev transfers an Agentless-based skill prior to SWE-Agent
through lightweight SFT, without additional multi-turn
RL~\citep{kimidev}. For RL-based agents, \syl{feedback} remains
largely trajectory-level, \syl{with limited supervision}
for intermediate tool decisions.

\subsection{On-Policy Distillation for Agents}
Agentic distillation methods regulate teacher guidance across
multi-turn trajectories. SDAR gates an OPSD auxiliary loss with
detached token statistics while retaining verifier-driven RL;
StepOPSD redistributes preference credit over action-centered
steps using hindsight-enriched contexts~\citep{sdar,stepopsd}.
SOD reweights OPD by step-level divergence, while TCOD
progressively expands the exposed trajectory
depth~\citep{sod,tcod}.
\syl{Another line asks what the teacher should see: AR-OPD replaces full-reference imitation with anchored residual guidance, ATESD learns how much of the reference to reveal, and EviSD gives search and answer actions different evidence~\citep{aropd,atesd,evisd}. These methods keep the exposure fixed during training, whereas \method orders the evidence across stages.}

\subsection{Subtask-Oriented Software Repair}
Agentless organizes repair into localization, patch generation,
and validation, while AutoCodeRover iteratively retrieves program
structure before patching~\citep{agentless,autocoderover}.
SWE-Fixer follows a retrieve-then-edit workflow, and SoRFT makes
file-, function-, and line-level localization explicit before
editing~\citep{swefixer,sorft}.
For testing, SWT-Bench evaluates reproduction through generated
tests, while AssertFlip converts passing tests into fail-to-pass
reproduction tests~\citep{swtbench,assertflip}.
SWE-Swiss trains localization, repair, and test generation through
multi-task SFT followed by targeted RL~\citep{swe_swiss}.
These works make repair subgoals explicit \syl{and fix them into the inference pipeline or the training data. \method keeps the agent end-to-end and uses this structure only} to align privileged teacher evidence with subtask-specific
response tokens and progressively schedule distillation\syl{. Curricula for LLM reinforcement learning usually order problems by difficulty~\citep{ladder}; \method keeps each task fixed and orders the evidence that the teacher sees.}

\section{Method}

\begin{figure*}[t]
  \centering
  \includegraphics[width=\linewidth]{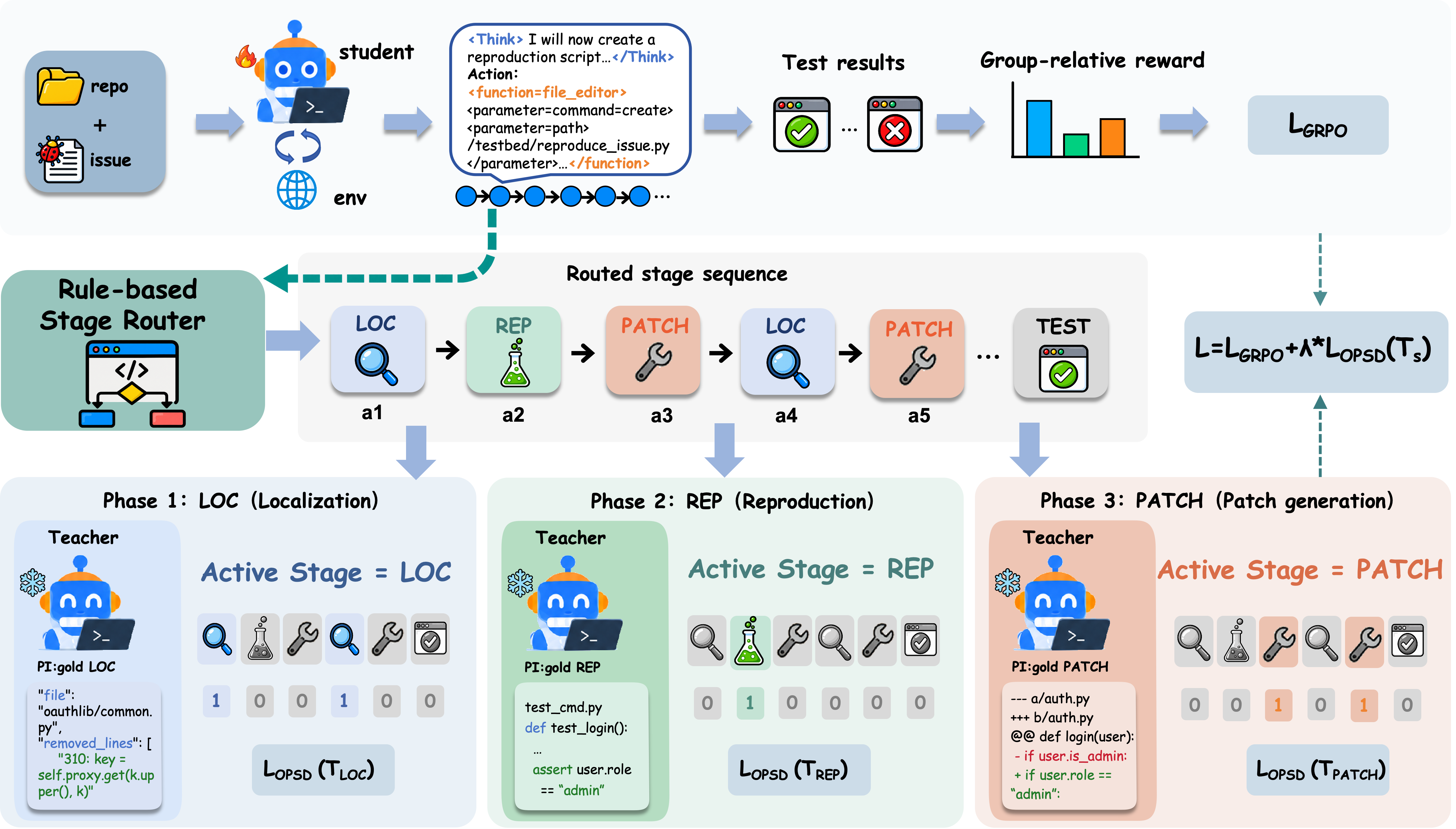}
  \caption{\textbf{Stage-routed privileged distillation for SWE.} \syl{A rule-based router labels each step of an end-to-end rollout as LOC, REP, or PATCH. In each phase, a frozen teacher receives one stage's evidence and distills only that stage's tokens, while GRPO optimizes the full trajectory.}}
  \label{fig:overview}
\end{figure*}

\syl{Following the analysis in Section~\ref{sec:intro} and} the staged organization of Agentless~\citep{agentless}, we organize
end-to-end software repair into three training-time subtasks: fault localization
(\textsc{LOC}), failure reproduction (\textsc{REP}), and patch generation
(\textsc{PATCH}). We collect a distinct source of task-specific privileged
evidence for each subtask. \method then
trains a single policy in the end-to-end repair setting while progressively
targeting one subtask at a time with its corresponding privileged evidence \syl{(Figure~\ref{fig:overview})}.
To target a subtask within an end-to-end rollout, we first assign each
assistant step a repair-stage label using its tool call. At each training
iteration, the schedule selects an active stage; its evidence conditions the
teacher, and only response tokens labeled with that stage receive the OPSD
loss. GRPO continues to optimize all valid response tokens, so stage-specific
supervision is integrated into full-trajectory learning.

\subsection{Rule-Based Stage Routing}
\label{sec:stage-routing}

\method uses a deterministic tool-level state machine to assign each generated
step one of \textsc{LOC}, \textsc{REP}, or \textsc{PATCH}, in that order,
while allowing state-dependent backtracking. The router labels response tokens
and routes only the OPSD teacher context; it does not constrain tool choices,
rollouts, or rewards. Marker rules, backtracking, malformed-tool handling, and
token alignment are given in Appendix~\ref{app:stage-annotation}. \syl{On held-out stage-labeled steps, the router agrees with the reference labels on 92.0\% of steps; a neural router reaches higher agreement but trains a worse agent (Section~\ref{sec:ablation-pspd}).}

\subsection{Stage-Specific Privileged Evidence}
\label{sec:privileged-evidence}

We construct three evidence views from each training instance's verified
repair metadata. We use only factual content from the repository, tests, and
verified repair; collecting this evidence introduces no knowledge generated
by external large language models. \syl{Each view serves} one subgoal:
\begin{itemize}[itemsep=1pt,topsep=2pt,parsep=0pt]
    \item \textbf{Localization evidence.} We extract changed non-test file
    paths and the old-side hunk ranges from the verified repair, together with
    the corresponding pre-patch source lines. These clues identify where to
    investigate without revealing the replacement code.
    \item \textbf{Reproduction evidence.} We collect fail-to-pass test
    identifiers and names, and include reference test code and candidate test
    changes when available. This evidence describes the failing behavior and
    provides a starting point for constructing a reproduction script.
    \item \textbf{Patch evidence.} We provide the verified \syl{repair diff, which specifies the fix.}
\end{itemize}
The evidence fields are stored with each instance, but teacher prompts are
rendered at training time. All three teacher variants share the student's
system instructions and end-to-end task prompt; only an appended evidence
block differs. Appendix~\ref{app:prompts} shows a routed trajectory, the
evidence fields for each stage, and their token cost. The active training
phase selects which evidence view is used
for distillation.

\subsection{Full-Trajectory GRPO Objective}
\label{sec:full-trajectory-rl}

We denote the standard GRPO objective~\citep{deepseekmath} for rollout $g$
at training iteration $u$ by $\mathcal L_{\mathrm{GRPO}}^{u,g}(\theta)$.
It acts on all valid model-generated response tokens throughout the
end-to-end trajectory, regardless of the active training stage. The stage
schedule changes only the auxiliary OPSD term defined below; the GRPO loss
and its token coverage remain unchanged. In particular, stage labels are
annotations for distillation, not early-stopping conditions: a rollout is
never truncated at a stage boundary and always runs to normal termination
or the configured resource limit.

\subsection{Stage-Conditioned OPSD and Joint Objective}
\label{sec:stage-conditioned-teacher}

For a fixed training iteration $u$ and rollout $g$, we suppress
$(u,g)$ in token-level quantities and per-rollout losses.
The active training stage $a(u)$, defined in
Section~\ref{sec:progressive-privilege}, selects the teacher's
privileged context, so $c_t=c_{a(u)}$ throughout the rollout.
The evidence views are described in
Section~\ref{sec:privileged-evidence} and
Appendix~\ref{app:prompts}.
Let $z_t\in\{\mathrm{LOC},\mathrm{REP},\mathrm{PATCH}\}$ denote
the stage label at token position $t$, assigned by the step-level
router in Section~\ref{sec:stage-routing}.

The teacher $q_{\phi_0}$ is initialized from the student's base
checkpoint and remains frozen throughout training.
Let $h_t$ denote the student-sampled interaction prefix before
token $t$, including earlier tool observations.
We abbreviate the student and teacher distributions as
$p_t(v)=p_\theta(v\mid h_t)$ and
$q_t(v)=q_{\phi_0}(v\mid h_t,c_t)$, respectively,
with the dependence of $p_t$ on $\theta$ implicit.
The teacher selects its $K$ highest-probability vocabulary tokens:
\begin{equation}
\mathcal K_t=\operatorname{TopK}_{v}\,q_t(v).
\end{equation}
Its distribution is then normalized over this set:
\begin{equation}
\widetilde q_t(v)
=\frac{q_t(v)}{\sum_{w\in\mathcal K_t}q_t(w)},
\qquad v\in\mathcal K_t.
\label{eq:teacher-topk-renorm}
\end{equation}
The student probabilities retain their full-vocabulary
normalization. We use the following stop-gradient
top-$K$ surrogate:
\begin{equation}
\begin{aligned}
d_t(\theta)
&=\sum_{v\in\mathcal K_t}p_t(v)\\
&\quad{}\times\bigl[
\operatorname{sg}(\log p_t(v))
-\operatorname{sg}(\log\widetilde q_t(v))
\bigr].
\end{aligned}
\label{eq:detached-topk-rkl}
\end{equation}
Here, $\operatorname{sg}$ denotes stop-gradient: both log terms
are detached, while gradients flow through the outer student
probability factor $p_t(v)$.
Implementation details are provided in
Appendix~\ref{app:topk-implementation}.

To restrict distillation to the active subtask, let
$m_t^{\mathrm{resp}}$ equal one at valid model-generated
response positions and zero elsewhere. The OPSD mask and
normalizer are
\begin{equation}
\begin{aligned}
m_t^{\mathrm{OPSD}}
&=m_t^{\mathrm{resp}}\,\mathbf{1}\{z_t=a(u)\},\\
N^{\mathrm{OPSD}}
&=\max\bigl\{1,\textstyle\sum_t m_t^{\mathrm{OPSD}}\bigr\}.
\end{aligned}
\label{eq:stage-opsd-mask}
\end{equation}
The per-rollout distillation loss and joint objective are
\begin{equation}
\begin{aligned}
\mathcal L_{\mathrm{OPSD}}(\theta)
&=\frac{1}{N^{\mathrm{OPSD}}}
\sum_t m_t^{\mathrm{OPSD}}d_t(\theta),\\
\mathcal L(\theta)
&=\mathcal L_{\mathrm{GRPO}}(\theta)
+\lambda\mathcal L_{\mathrm{OPSD}}(\theta),
\end{aligned}
\label{eq:pspd-total}
\end{equation}
where $\lambda$ controls the distillation weight.
GRPO remains active on all valid response tokens, including
those outside the active subtask. 

\subsection{Progressive Privilege Scheduling}
\label{sec:progressive-privilege}

\method applies privileged stages in the fixed order
$\mathrm{LOC}\rightarrow\mathrm{REP}\rightarrow\mathrm{PATCH}$, with each
phase continuing from the preceding checkpoint. The active stage $a(u)$
selects both the teacher's privileged evidence and the response tokens used
for OPSD via Eq.~(\ref{eq:stage-opsd-mask}). \syl{The order follows how repair uses the evidence. On-policy distillation supervises only the states that the student reaches, so patch evidence helps only once the student can find the code to change, and a reproduction script is what the student later runs to check its patch. Section~\ref{sec:ablation-pspd} compares alternative orders.} The formal schedule,
phase budgets, and training and inference details are given in
Appendix~\ref{app:progressive-schedule}.

\section{Experiments}


\subsection{Experimental Setup}
\label{sec:setup-pspd}

\begin{table}[t]
    \centering
    \small
    \setlength{\tabcolsep}{3pt}
    \renewcommand{\arraystretch}{1.05}

    \resizebox{\linewidth}{!}{%
    \begin{tabular}{@{}lrr@{}}
        \toprule
        Source corpus & Instances & Share (\%) \\
        \midrule
        SWE-smith~\citep{swesmith}   & 500       & 43.5  \\
        SWE-ReBench~\citep{swerebench} & 431       & 37.5  \\
        R2E-Gym~\citep{r2egym}     & 155       & 13.5  \\
        SWE-Gym~\citep{swegym}     & 63        & 5.5   \\
        \midrule
        Total       & $1{,}149$ & 100.0 \\
        \bottomrule
    \end{tabular}}
  \vspace{2pt}
  \caption{\textbf{Training data composition.} The collection contains
    1,149 issues from four software-repair corpora.}
  \label{tab:training-data-pspd}
\end{table}

\textbf{Benchmark and protocol.} We evaluate on SWE-bench Verified
, which contains $500$ human-validated repository issues, and
report pass@1\citep{swebench,swebenchverified}. An issue is counted as solved only when the submitted patch
passes all fail-to-pass and pass-to-pass unit tests. The primary RL
comparisons use the same issues, R2E-Gym~\citep{r2egym} agent scaffold, sandbox, and
evaluation harness.

\textbf{Model and training.} We use the Qwen3-4B-Instruct-2507 policy
backbone~\citep{qwen3}. There is no additional task-specific supervised
fine-tuning before RL. \method
uses a frozen teacher initialized from the same base checkpoint as the
student. Each training iteration uses a batch size of $16$ tasks and a group
size of $4$ rollouts per task, yielding $64$ rollouts per iteration. We use a
teacher top-set of $K=16$ tokens, an entropy-bonus coefficient of $0.005$, and
$\lambda=0.15$ for the OPSD term. The training collection contains $1{,}149$
issues from four software-repair corpora; their composition is summarized in
Table~\ref{tab:training-data-pspd}.

\textbf{Progressive training.} \method uses the fixed
LOC$\rightarrow$REP$\rightarrow$PATCH schedule with $30$, $20$, and $20$
global iterations, respectively. Every rollout remains end to end through
normal termination or the configured resource limit. The rule-based router
assigns stages to assistant steps; OPSD selects only the response tokens
assigned to the active phase, while GRPO retains all valid response tokens.
Detailed scheduling and training configuration are given in
Appendix~\ref{app:progressive-schedule}.

\textbf{Comparison scope.} Our primary baselines are GRPO and
GRPO+OPSD, which exposes the reference patch \syl{at every step} without stage-specific routing.
We additionally include SDAR\syl{~\citep{sdar}, a gated OPSD method for agents, with the same patch evidence. All methods start from the same base model and are evaluated with the same scaffold and harness; training curves cover a common 70-iteration horizon (Appendix~\ref{app:training-curves}).}

\subsection{Main Results}
\label{sec:main-pspd}

\syl{\paragraph{\method improves end-to-end repair on both scaffolds.} On R2E-Gym, \method raises pass@1 from 9.6\% to 16.4\%, compared with 10.4\% for GRPO and 12.4\% for GRPO+OPSD (Table~\ref{tab:main-pspd}), without task-specific supervised fine-tuning. It also uses fewer steps than the base model (19.3 vs.\ 26.5) and completes 98.6\% of its trajectories, whereas GRPO completes only 36.0\% before the resource limit. With the same patch evidence, SDAR falls below the base model (4.6\%).}

\begin{table*}[!t]
  \centering

  \sweTabMainSetup

  \begin{tabularx}{\linewidth}{
    >{\raggedright\arraybackslash}p{0.34\linewidth}
    >{\centering\arraybackslash}p{0.10\linewidth}
    *{3}{>{\centering\arraybackslash}X}
  }
    \toprule
    \textbf{Configuration}
      & \textbf{Training}
      & \textbf{pass@1 (\%)} $\uparrow$
      & \textbf{Avg.\ steps} $\downarrow$
      & \textbf{\texttt{env\_done} (\%)} $\uparrow$ \\
    \midrule

    \multicolumn{5}{l}{\textbf{R2E-Gym}} \\
    \addlinespace[2pt]

    \rowcolor{swepreviewbasebg}
    Qwen3-4B-Instruct-2507 (Base) & ---
      & \sweTabCell{9.6}{}
      & \sweTabCell{26.5}{}
      & \sweTabCell{84.2}{} \\

    GRPO & RL
      & \sweTabCell{10.4}{\sweTabDelta{sweTabGreen}{\uparrow}{+}{0.8}}
      & \sweTabCell{82.9}{\sweTabDelta{sweTabRed}{\uparrow}{+}{56.5}}
      & \sweTabCell{36.0}{\sweTabDelta{sweTabRed}{\downarrow}{-}{48.2}} \\

    GRPO+OPSD & RL
      & \sweTabCell{12.4}{\sweTabDelta{sweTabGreen}{\uparrow}{+}{2.8}}
      & \sweTabCell{39.7}{\sweTabDelta{sweTabRed}{\uparrow}{+}{13.3}}
      & \sweTabCell{67.4}{\sweTabDelta{sweTabRed}{\downarrow}{-}{16.8}} \\

    SDAR & RL
      & \sweTabCell{4.6}{\sweTabDelta{sweTabRed}{\downarrow}{-}{5.0}}
      & \sweTabCell{34.5}{\sweTabDelta{sweTabRed}{\uparrow}{+}{8.0}}
      & \sweTabCell{88.0}{\sweTabDelta{sweTabGreen}{\uparrow}{+}{3.8}} \\

    \textbf{\method} & RL
      & \sweTabCell{16.4}{\sweTabDelta{sweTabGreen}{\uparrow}{+}{6.8}}
      & \sweTabCell{19.3}{\sweTabDelta{sweTabGreen}{\downarrow}{-}{7.2}}
      & \sweTabCell{98.6}{\sweTabDelta{sweTabGreen}{\uparrow}{+}{14.4}} \\

    \addlinespace[4pt]
    \midrule
    \multicolumn{5}{l}{\textbf{OpenHands}} \\
    \addlinespace[2pt]

    \rowcolor{swepreviewbasebg}
    Qwen3-4B-Instruct-2507 (Base) & ---
      & \sweTabCell{13.0}{}
      & \sweTabCell{40.8}{}
      & \sweTabCell{79.2}{} \\

    GRPO & RL
      & \sweTabCell{11.2}{\sweTabDelta{sweTabRed}{\downarrow}{-}{1.8}}
      & \sweTabCell{56.2}{\sweTabDelta{sweTabRed}{\uparrow}{+}{15.4}}
      & \sweTabCell{78.6}{\sweTabDelta{sweTabRed}{\downarrow}{-}{0.6}} \\

    GRPO+OPSD & RL
      & \sweTabCell{12.4}{\sweTabDelta{sweTabRed}{\downarrow}{-}{0.6}}
      & \sweTabCell{15.4}{\sweTabDelta{sweTabGreen}{\downarrow}{-}{25.4}}
      & \sweTabCell{99.0}{\sweTabDelta{sweTabGreen}{\uparrow}{+}{19.8}} \\

    SDAR & RL
      & \sweTabCell{6.8}{\sweTabDelta{sweTabRed}{\downarrow}{-}{6.2}}
      & \sweTabCell{37.4}{\sweTabDelta{sweTabGreen}{\downarrow}{-}{3.4}}
      & \sweTabCell{92.8}{\sweTabDelta{sweTabGreen}{\uparrow}{+}{13.6}} \\

    \textbf{\method} & RL
      & \sweTabCell{17.6}{\sweTabDelta{sweTabGreen}{\uparrow}{+}{4.6}}
      & \sweTabCell{16.6}{\sweTabDelta{sweTabGreen}{\downarrow}{-}{24.2}}
      & \sweTabCell{99.4}{\sweTabDelta{sweTabGreen}{\uparrow}{+}{20.2}} \\

    \bottomrule
  \end{tabularx}

  \par\smallskip
  \begin{minipage}{\linewidth}
    \footnotesize
    \textit{Note.} Colored arrows show absolute changes from the Qwen3 base
    within each scaffold: percentage points for pass@1 and \texttt{env\_done},
    and steps for Avg. steps. For Avg. steps, a downward arrow is favorable.
    All OpenHands rows use the common evaluation protocol in
    Appendix~\ref{app:openhands-eval}.
  \end{minipage}
  \vspace{2pt}
  \caption{\textbf{SWE-bench Verified results.}
  We report results for both the R2E-Gym and OpenHands~\citep{openhands}
  scaffolds. Each configuration is evaluated in our environment with the
  scaffold listed in the table. All reported baselines and \method variants
  are evaluated in the same environment and protocol.}
  \label{tab:main-pspd}
\end{table*}

\syl{On the OpenHands scaffold, the three baselines stay at or below the base model:} the base policy reaches $13.0\%$ pass@1\syl{, and} GRPO,
GRPO+OPSD, and SDAR reach $11.2\%$, $12.4\%$, and $6.8\%$,
respectively, whereas \method from Base reaches $17.6\%$ with
\syl{$16.6$ steps and $99.4\%$ completion}.

\begin{figure}[!t]
  \centering
  \includegraphics[width=\columnwidth]{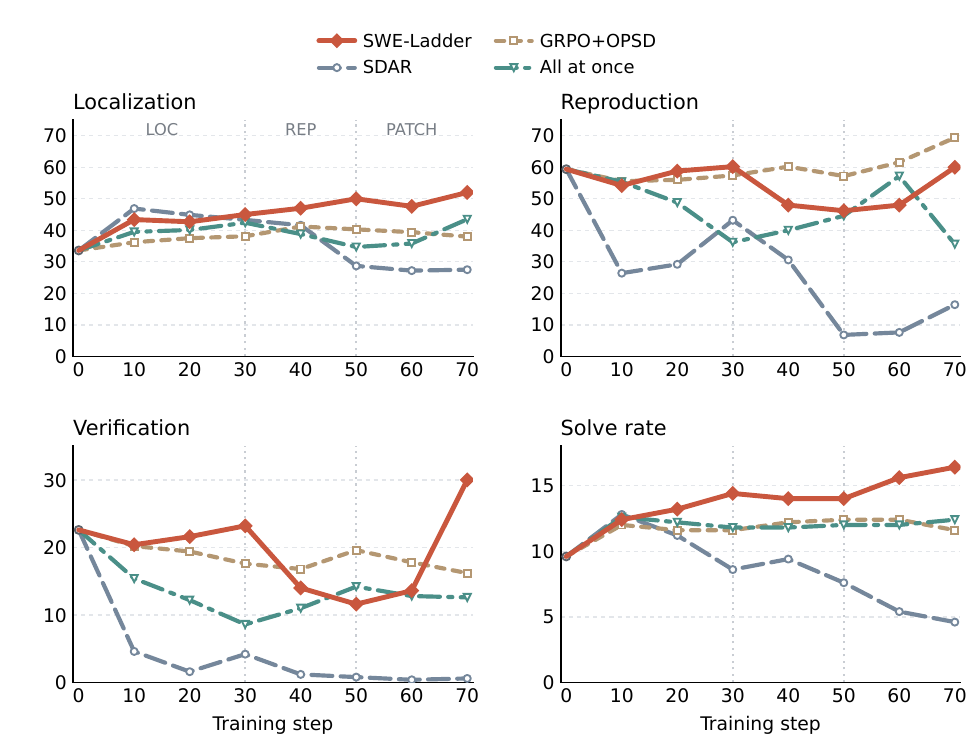}
  \caption{\textbf{Four-view training dynamics on SWE-bench Verified.}
  \syl{Localization, reproduction, verification, and pass@1, evaluated every 10 training iterations,} for SWE-Ladder, SDAR, GRPO+OPSD,
  and the \syl{all-stages-at-once} variant. \syl{Vertical guides mark the LOC, REP, and PATCH phases of \method.}}
  \label{fig:solve-rate-comparison}
\end{figure}

\syl{Figure~\ref{fig:solve-rate-comparison} tracks the four metrics during training. The localization of \method rises from 33.6\% to 52.0\%. Its reproduction and verification dip during the REP phase and recover during the PATCH phase, ending above the base model. GRPO+OPSD raises reproduction but lowers verification, SDAR and the all-stages-at-once variant fall below the base model on both, and only \method ends above it on all four metrics.}

\subsection{Training Dynamics}
\label{sec:training-dynamics}

\begin{figure}[!t]
  \centering
  \includegraphics[width=\columnwidth]{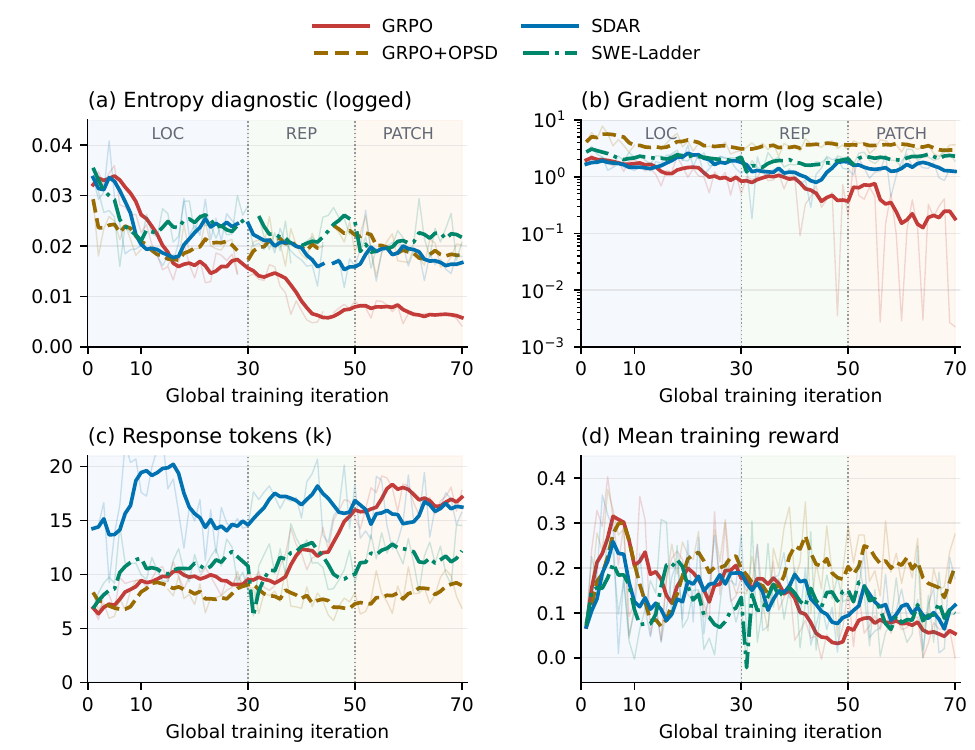}
  \caption{\textbf{Training-dynamics diagnostics.}
  \syl{Entropy, gradient norm, response length, and shaped training reward of GRPO, GRPO+OPSD, SDAR, and \method over 70 iterations. Shaded regions mark the LOC, REP, and PATCH phases of \method.}}
  \label{fig:dynamics-audited}
\end{figure}

To investigate training stability, we compare \method with
GRPO, GRPO+OPSD, and SDAR. Figure~\ref{fig:dynamics-audited}
tracks policy entropy, gradient norm, response length, and shaped training
reward across individual training runs
(Appendix~\ref{app:dynamics-pspd}).

GRPO exhibits substantial deterioration in its training
dynamics. Comparing the first and last ten iterations,
its mean entropy decreases from $0.0300$ to $0.0061$,
while its gradient norm drops from $1.820$ to $0.187$.
Meanwhile, the average response length nearly doubles
from $8.29$k to $16.73$k tokens, \syl{and training reward declines}.

In contrast, \method exhibits more consistent optimization
dynamics, with relatively stable entropy
($0.0251 \rightarrow 0.0221$) and gradient norm
($2.353 \rightarrow 2.397$), alongside a modest increase
in response length ($10.09$k to $11.62$k tokens).
Although its training reward remains variable,
\method does not exhibit the pronounced diagnostic
deterioration observed under GRPO.

GRPO+OPSD and SDAR also retain optimization
signals late in training. Thus, maintaining entropy or
gradient magnitude alone does not explain \method's
superior held-out performance. These observations suggest
more consistent optimization behavior under the tested
configuration, without establishing a general stability
guarantee.

\subsection{Ablation Study}
\label{sec:ablation-pspd}

\begin{figure*}[!t]
  \centering
  \includegraphics[width=\textwidth]{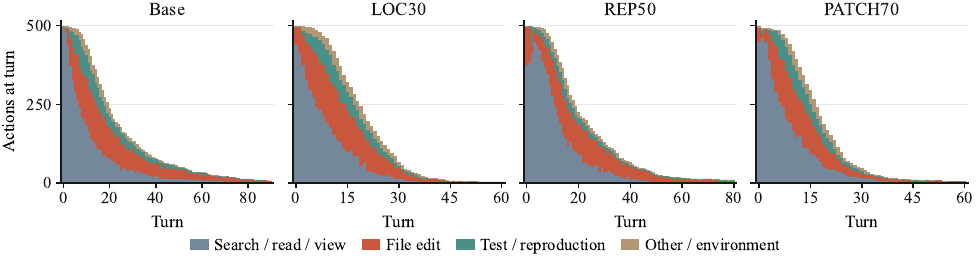}
  \caption{\textbf{\syl{Action counts per turn across checkpoints.}} \syl{For the base model, \texttt{LOC30}, \texttt{REP50}, and \texttt{PATCH70}, each panel stacks the number of trajectories taking a search/read/view, file-edit, test/reproduction, or other action at each turn on SWE-bench Verified ($n=500$).}}
  \label{fig:checkpoint-action-distribution}
\end{figure*}

\begin{figure}[!b]
  \centering
  \includegraphics[width=\columnwidth]{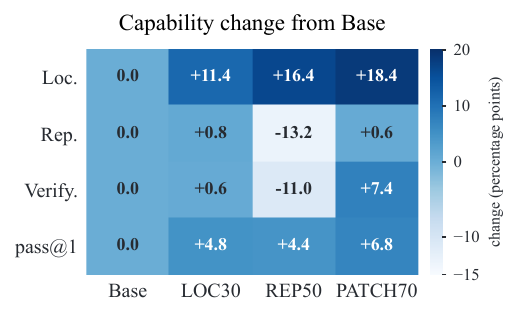}
  \caption{\textbf{Capability changes across checkpoints.}
  Cells show \syl{changes (\%)} from \texttt{Base} for
  localization (Loc.), reproduction (Rep.), verification (Verify.), and
  pass@1 at \texttt{Base}, \texttt{LOC30}, \texttt{REP50}, and
  \texttt{PATCH70}.
  Darker blue indicates a larger signed change.}
  \label{fig:checkpoint-capability-heatmap}
\end{figure}

We organize the ablations around three questions: whether all three training
stages contribute, whether a deterministic router is preferable to a neural
router for training, and whether the privileged information is matched to the
stage whose response tokens it supervises.

\paragraph{All three stages contribute to capability gains.}
We remove each stage from the progressive schedule in turn and compare with
a variant that trains all three stages jointly. \syl{In this variant},
the LOC, REP, and PATCH teacher contexts are available concurrently, and
each routed response-token subset is distilled from its corresponding context;
it does not use the progressive continuation schedule. \syl{We refer to it as \emph{all stages at once}.} The variants share the
Qwen3-4B-Instruct-2507 base and are evaluated with the same R2E-Gym scaffold
and protocol.

\begin{table}[t]
  \centering

  \sweTabAblationSetup

  \begin{tabularx}{\linewidth}{
    >{\raggedright\arraybackslash}p{0.25\linewidth}
    *{4}{>{\centering\arraybackslash}X}
  }
    \toprule
    \textbf{Variant}
      & \textbf{Loc.}
      & \textbf{Rep.}
      & \textbf{Verify.}
      & \textbf{pass@1} \\
    \midrule

    \rowcolor{swepreviewbasebg}
    Base
      & \sweTabCell{33.6}{}
      & \sweTabCell{59.4}{}
      & \sweTabCell{22.6}{}
      & \sweTabCell{9.6}{} \\

    \leavevmode\syl{All at once}
      & \sweTabCell{43.5}{\sweTabDelta{sweTabGreen}{\uparrow}{+}{9.9}}
      & \sweTabCell{35.6}{\sweTabDelta{sweTabRed}{\downarrow}{-}{23.8}}
      & \sweTabCell{12.6}{\sweTabDelta{sweTabRed}{\downarrow}{-}{10.0}}
      & \sweTabCell{12.4}{\sweTabDelta{sweTabGreen}{\uparrow}{+}{2.8}} \\

    REP + PATCH
      & \sweTabCell{39.9}{\sweTabDelta{sweTabGreen}{\uparrow}{+}{6.3}}
      & \sweTabCell{29.6}{\sweTabDelta{sweTabRed}{\downarrow}{-}{29.8}}
      & \sweTabCell{11.8}{\sweTabDelta{sweTabRed}{\downarrow}{-}{10.8}}
      & \sweTabCell{11.2}{\sweTabDelta{sweTabGreen}{\uparrow}{+}{1.6}} \\

    LOC + PATCH
      & \sweTabCell{42.4}{\sweTabDelta{sweTabGreen}{\uparrow}{+}{8.8}}
      & \sweTabCell{37.6}{\sweTabDelta{sweTabRed}{\downarrow}{-}{21.8}}
      & \sweTabCell{8.2}{\sweTabDelta{sweTabRed}{\downarrow}{-}{14.4}}
      & \sweTabCell{13.2}{\sweTabDelta{sweTabGreen}{\uparrow}{+}{3.6}} \\

    LOC + REP
      & \sweTabCell{50.0}{\sweTabDelta{sweTabGreen}{\uparrow}{+}{16.4}}
      & \sweTabCell{46.2}{\sweTabDelta{sweTabRed}{\downarrow}{-}{13.2}}
      & \sweTabCell{11.6}{\sweTabDelta{sweTabRed}{\downarrow}{-}{11.0}}
      & \sweTabCell{14.0}{\sweTabDelta{sweTabGreen}{\uparrow}{+}{4.4}} \\

    \textbf{\method}
      & \sweTabCell{52.0}{\sweTabDelta{sweTabGreen}{\uparrow}{+}{18.4}}
      & \sweTabCell{60.0}{\sweTabDelta{sweTabGreen}{\uparrow}{+}{0.6}}
      & \sweTabCell{30.0}{\sweTabDelta{sweTabGreen}{\uparrow}{+}{7.4}}
      & \sweTabCell{16.4}{\sweTabDelta{sweTabGreen}{\uparrow}{+}{6.8}} \\

    \bottomrule
  \end{tabularx}
  \vspace{2pt}
  \caption{\textbf{Stage ablation on SWE-bench Verified.}
  Qwen3-4B-Instruct-2507 with R2E-Gym, $n=500$.
  We report Loc., Rep., Verify., and pass@1; arrows indicate
  changes relative to Base. \syl{``All at once''} applies the
  LOC, REP, and PATCH teacher contexts concurrently, with each routed
  response-token subset distilled from its corresponding context.}
  \label{tab:stage-ablation-pspd}
\end{table}

Table~\ref{tab:stage-ablation-pspd} shows that removing LOC, REP, or PATCH
reduces pass@1 from $16.4\%$ to $11.2\%$, $13.2\%$, or $14.0\%$,
respectively\syl{, and every variant falls below the base model on reproduction and verification. Removing LOC causes the largest drop, consistent with localization being the stage on which the others depend. Training all stages at once reaches 12.4\% and also loses reproduction (35.6\%) and verification (12.6\%), so routing the evidence without the order recovers neither pass@1 nor the intermediate behaviors. The REP phase does not raise the behavior metrics by itself (Section~\ref{sec:checkpoint-behavior}), but removing it still lowers the final pass@1.}

\paragraph{Stage order matters.}
We examine whether the order of privileged supervision affects
training outcomes by comparing the proposed
LOC$\rightarrow$REP$\rightarrow$PATCH schedule with
PATCH$\rightarrow$REP$\rightarrow$LOC and
REP$\rightarrow$LOC$\rightarrow$PATCH.
All schedules allocate 30 iterations to LOC and 20 each to REP
and PATCH\syl{.}

As shown in Table~\ref{tab:stage-order}, the proposed order
achieves $16.4\%$ pass@1, outperforming the reverse order
($11.4\%$) and the REP-first order ($12.6\%$)\syl{.} It also requires fewer
interaction steps and achieves a higher environment completion
rate. Notably, REP$\rightarrow$LOC$\rightarrow$PATCH retains
PATCH as the final phase but still underperforms the proposed
schedule, highlighting the importance of the preceding stage
order. These results support introducing localization before
reproduction, followed by patch generation.
\begin{table}[t]
  \centering

  \sweTabOrderSetup

  \begin{tabularx}{\linewidth}{
    >{\raggedright\arraybackslash}p{0.34\linewidth}
    *{3}{>{\centering\arraybackslash}X}
  }
    \toprule
    \textbf{Training order}
      & \textbf{Solve (\%)}
      & \textbf{Avg. steps}
      & \textbf{env\_done (\%)} \\
    \midrule

    \rowcolor{swepreviewbasebg}
    LOC$\to$REP$\to$PATCH
      & \sweTabCell{16.4}{}
      & \sweTabCell{19.3}{}
      & \sweTabCell{98.6}{} \\

    PATCH$\to$REP$\to$LOC
      & \sweTabCell{11.4}{
          \sweTabDelta{sweTabRed}{\downarrow}{-}{5.0}}
      & \sweTabCell{27.6}{
          \sweTabDelta{sweTabRed}{\uparrow}{+}{8.3}}
      & \sweTabCell{95.8}{
          \sweTabDelta{sweTabRed}{\downarrow}{-}{2.8}} \\

    REP$\to$LOC$\to$PATCH
      & \sweTabCell{12.6}{
          \sweTabDelta{sweTabRed}{\downarrow}{-}{3.8}}
      & \sweTabCell{28.5}{
          \sweTabDelta{sweTabRed}{\uparrow}{+}{9.2}}
      & \sweTabCell{96.0}{
          \sweTabDelta{sweTabRed}{\downarrow}{-}{2.6}} \\

    \bottomrule
  \end{tabularx}
  \vspace{2pt}
  \caption{
    \textbf{Effect of training-stage order.}
    All schedules use 70 training iterations:
    30 for LOC and 20 each for REP and PATCH.
    The blue row is the reference.
    Arrows show changes in percentage points for Solve and
    env\_done, and in steps for Avg. steps.
  }
  \label{tab:stage-order}
\end{table}

\paragraph{Rule-based routing is more effective for training.}
The neural router achieves higher agreement with the stage
annotations than the rule-based router
($98.63\%$ vs.\ $92.0\%$), yet yields lower downstream
pass@1. This gap may reflect more fragmented stage assignments
under independent step prediction, compared with stateful
rule-based routing. Annotation agreement alone therefore
does not determine training utility. Details are provided
in Appendix~\ref{app:neural-router}.

\paragraph{Privilege should match the target stage.}
We compare matched and swapped LOC/PATCH evidence under a
30-iteration budget (Table~\ref{tab:privilege-alignment-pspd}).
For LOC targets, matched LOC PI achieves $14.4\%$ pass@1,
compared with $13.6\%$ when using PATCH PI, indicating that patch
evidence retains some localization cues. For PATCH targets, matched
PATCH PI reaches $13.2\%$, compared with $11.8\%$ for LOC PI,
showing that localization evidence alone exposes too little
information for patch generation. Although the matched PATCH
condition uses longer trajectories and has lower
\texttt{env\_done} than the mismatched condition, it achieves higher
repair success, indicating that shorter interactions alone do not
imply better repairs.

\begin{table}[t]
  \centering

  \sweTabAlignSetup

  \begin{tabularx}{\linewidth}{
    >{\centering\arraybackslash}p{0.10\linewidth}
    >{\centering\arraybackslash}p{0.23\linewidth}
    *{3}{>{\centering\arraybackslash}X}
  }
    \toprule
    \textbf{Target}
      & \textbf{PI $\to$ target}
      & \textbf{Solve (\%)}
      & \textbf{Avg. steps}
      & \textbf{env\_done (\%)} \\
    \midrule

    \rowcolor{swepreviewbasebg}
    \multirow{2}{*}{LOC}
      & LOC$\to$LOC
      & \sweTabCell{14.4}{}
      & \sweTabCell{21.0}{}
      & \sweTabCell{99.0}{} \\

      & PATCH$\to$LOC
      & \sweTabCell{13.6}{
          \sweTabDelta{sweTabRed}{\downarrow}{-}{0.8}}
      & \sweTabCell{15.9}{
          \sweTabDelta{sweTabGreen}{\downarrow}{-}{5.1}}
      & \sweTabCell{98.8}{
          \sweTabDelta{sweTabRed}{\downarrow}{-}{0.2}} \\

    \addlinespace[2pt]

    \rowcolor{swepreviewbasebg}
    \multirow{2}{*}{PATCH}
      & PATCH$\to$PATCH
      & \sweTabCell{13.2}{}
      & \sweTabCell{34.0}{}
      & \sweTabCell{82.8}{} \\

      & LOC$\to$PATCH
      & \sweTabCell{11.8}{
          \sweTabDelta{sweTabRed}{\downarrow}{-}{1.4}}
      & \sweTabCell{20.2}{
          \sweTabDelta{sweTabGreen}{\downarrow}{-}{13.8}}
      & \sweTabCell{99.0}{
          \sweTabDelta{sweTabGreen}{\uparrow}{+}{16.2}} \\

    \bottomrule
  \end{tabularx}
  \vspace{2pt}
  \caption{
    \textbf{Privilege--target stage alignment.}
    All variants use a 30-step training budget.
    Rows are grouped by target stage; the aligned row in
    each group is the reference, and arrows show absolute
    changes for the mismatched PI condition.
  }
  \label{tab:privilege-alignment-pspd}
\end{table}

\subsection{Analysis}
\label{sec:checkpoint-behavior}
To examine how behavior changes along the progressive sequence, we use four
held-out evaluation snapshots and refer to them consistently as
\texttt{Base}, \texttt{LOC30}, \texttt{REP50}, and \texttt{PATCH70}.
Figure~\ref{fig:checkpoint-action-distribution} provides a complementary view
of the same checkpoints from their audited trajectories. The stacks count
search/read/view, file-edit, test/reproduction, and other/environment actions
at each event turn; they are counts rather than normalized rates. The action
taxonomy \syl{comes from} the collected records and \syl{is not}
a stage-router annotation.

\syl{Figure~\ref{fig:checkpoint-capability-heatmap} shows how the four metrics change along the progressive sequence, at the end of each phase (\texttt{LOC30}, \texttt{REP50}, and \texttt{PATCH70}). The LOC phase improves all four metrics. The REP phase further improves localization, but reproduction and verification fall below the base model. The PATCH phase recovers both, and \texttt{PATCH70} exceeds the base model on all four metrics.}

\FloatBarrier
\section{Conclusion}



\syl{We studied what a privileged teacher should see when training an end-to-end SWE agent. \method gives the teacher only the evidence for the current stage and introduces the stages in repair order. It improves pass@1 on SWE-bench Verified from 9.6\% to 16.4\% with R2E-Gym and from 13.0\% to 17.6\% with OpenHands, and ablations show that both stage-matched evidence and the stage order matter.}

\section*{Limitations}

The evidence is bounded to one Qwen3-4B policy, the SWE-bench Verified
benchmark, and two agent scaffolds.  Although the OpenHands results provide a
second scaffold check, the main ablations and the stage-wise diagnostics use
the R2E-Gym protocol.  We therefore do not establish transfer to larger
backbones, other repair benchmarks, or substantially different tool
interfaces.  The reported training curves and router comparison each follow
one checkpoint lineage without independent random-seed repeats or confidence
intervals, so the observed optimization trends and downstream differences
should be interpreted as configuration-specific evidence.

The method also depends on verified repair metadata to construct the three
privileged views and on a parser that maps tool calls to repair stages.  The
neural-router accuracy is agreement with Qwen3.8-Max annotations rather than
an independently adjudicated semantic label, while the rule-based router can
miss unfamiliar tools or atypical backtracking patterns. Future work should test neural and
hybrid routers across models and benchmarks, repeat the training with
independent seeds, and quantify the cost and robustness of evidence
construction.



\section*{\syl{Ethics Statement}}

This work uses publicly available SWE-bench Verified issues, open-weight Qwen3
models, the hosted Qwen3.8-Max model for rollout stage annotation, and public
R2E-Gym and OpenHands scaffolds. We do not recruit human
participants or collect new personal data. Training and evaluation are
conducted in isolated benchmark sandboxes and do not take actions on external
systems.
Privileged repair evidence is used only as training-time context and is absent
at inference. Deploying agents trained with \method beyond these benchmark
sandboxes would require additional safety validation, permission controls, and
human oversight.



\bibliography{references}

\clearpage
\appendix
\section{Implementation Details for End-to-End Privileged Distillation}

\subsection{Stage-wise Capability Evaluation}
\label{app:stage-capability}

Figure~\ref{fig:stage_capability} reports four complementary
capability measurements on the same 500 held-out SWE-bench Verified tasks.
For each method, we use its selected evaluation model and evaluate one agent
trajectory per task under the common evaluation scaffold.  The first three
measurements are behavioral diagnostics extracted from the trajectory; they are
not additional verifier rewards.

\begin{itemize}
\item \textbf{Localization (gold-file recall).}  Let $F$ be the set of
      non-test files changed by the reference patch and $E$ the set of files
      changed by the agent's final patch.  We report
      $|F\cap E|/|F|$, averaged over tasks with a nonempty reference set.
      This measures whether the agent edits the files implicated by the issue,
      rather than whether the final patch is correct.
\item \textbf{Reproduction (ran failing repro).}  We count a trajectory as
      reproducing the issue when the agent creates a reproduction script,
      executes it, and observes a failure signal such as a traceback,
      assertion failure, failed test, exception, or nonzero command result.
\item \textbf{Verification (re-ran repro, passed).}  After the last source
      edit, we check whether the agent re-executes a previously created
      reproduction script without a failure signal.  This captures an explicit
      post-edit verification action and does not require the final patch to
      pass the hidden benchmark tests.
\item \textbf{Solve (pass@1).}  We use the official SWE-bench verifier result:
      a trajectory counts as solved when its submitted patch satisfies the
      designated fail-to-pass tests without introducing a regression.
\end{itemize}

We report each quantity as a percentage, but the aggregation differs by
metric: localization is the macro-average of task-level recalls over tasks
with a nonempty reference set, whereas reproduction, verification, and solve
are trajectory-level percentages.  The diagnostic definitions use only the
agent trajectory, reference patch file paths, and verifier outcome; they do
not expose privileged teacher evidence to the student.  Consequently, the
figure should be read as a stage-wise behavioral comparison: a method can
improve solve rate while becoming less reliable on an earlier subtask, which
is the failure mode the routed objective is designed to avoid.

\subsection{Stage Annotation and Token Alignment}
\label{app:stage-annotation}

The main method uses the deterministic rule-based router described in
Section~\ref{sec:stage-routing}. The parser consumes only the parsed tool
call: function name and arguments (the editor command and
\texttt{path}/\texttt{file\_path}, or bash \texttt{cmd}/\texttt{command}). It
accepts both standard and malformed inline parameter forms and supports the
r2egym and OpenHands tool dialects. Creating or editing
a file whose basename contains \texttt{repro} or \texttt{reproduce}, executing
the corresponding Python script, creating or editing a test file, or running
the project's test suite proposes \textsc{REP}. A test file is under
\texttt{/tests/} or \texttt{/test/}, or has basename \texttt{test\_*},
\texttt{*\_test.py}, \texttt{*\_tests.py}, \texttt{*tests.py},
\texttt{edge\_case*}, \texttt{conftest.py}, or contains \texttt{fixture}.
Creating or editing any other source file proposes \textsc{PATCH}. Search,
editor views, and read-only shell commands whose first token is in the
exploration set are exploration actions; all other tools emit no proposal.

A proposed stage advances the current state only if it is later in the order
\textsc{LOC}$\rightarrow$\textsc{REP}$\rightarrow$\textsc{PATCH}; other
actions normally inherit the current state, and \textsc{LOC}$\rightarrow$
\textsc{PATCH} can therefore skip \textsc{REP}. Backtracking is
state-dependent: in \textsc{REP}, three consecutive exploration actions
return the state to \textsc{LOC}; in \textsc{PATCH}, a single test run demotes
it to \textsc{REP}, while a single exploration action returns it to
\textsc{LOC}. These transitions are applied after the current step's proposal.

The state after processing a step labels that same step. The label is copied to
its assistant reasoning and action tokens and to the environment-token
positions concatenated into the response tensor. A separate response mask
excludes environment observations from optimization: GRPO uses all valid
model-generated response positions, while OPSD intersects this mask with the
active-stage indicator in Eq.~\eqref{eq:stage-opsd-mask}. The annotation does
not constrain tool choices, rollouts, or rewards, and it routes only the OPD
teacher context.

\subsection{Neural Router and Rule-Based Router Comparison}
\label{app:neural-router}

We also evaluate a lightweight neural router as an alternative to the rules,
rather than as a component of the main method. Qwen3.8-Max~\citep{qwen38max} labels assistant
steps from end-to-end Qwen3-4B-Instruct-2507
trajectories as \textsc{LOC}, \textsc{REP}, or \textsc{PATCH}, using the issue
excerpt together with action and observation snippets. The resulting data
contain 10,440 valid labels from 731 trajectories. We use a deterministic
trajectory-level 80/20 hash split, giving 8,179 training labels and 2,261
held-out labels. The router concatenates the parsed tool function name with
the lower-cased assistant action, applies a TF--IDF unigram/bigram transform
with 50,000 maximum features, and predicts the three labels with a
class-balanced linear classifier ($C=4$). Its held-out accuracy against
the Qwen3.8-Max annotations is 98.63\%. The deterministic rule-based router
obtains 92.0\% agreement on the same 2,261 held-out records. These figures measure
agreement with the annotator on held-out trajectories, not independently
adjudicated semantic accuracy or generalization to unseen issues. Qwen3.8-Max is used only to
produce these rollout annotations; it is not an OPSD teacher.

After each assistant message is generated, this classifier independently
predicts one canonical stage for the step, without temporal smoothing or a
monotonicity constraint. Consequently, small changes in an action or its
parsed tool call can change the predicted stage and fragment a trajectory into
isolated labels. The rule-based router instead uses an ordered state machine
with state-dependent backtracking; its assignments are consequently robust to
local action variation and tend to form contiguous stage blocks.
Because the assistant action text includes any reasoning
tokens present in that message, the classifier input includes that text as
well as the parsed function name; it does not use model hidden states. The
predicted label is copied to every assistant and environment-token position
generated by the step. If the classifier cannot be loaded or produces an
invalid label, the implementation falls back to the rule-based state machine.
Its annotation-agreement score and downstream solve rate assess different
properties: high agreement with the annotator does not by itself establish
better training supervision.

\begin{figure}[!htbp]
  \centering
  \includegraphics[width=\linewidth]{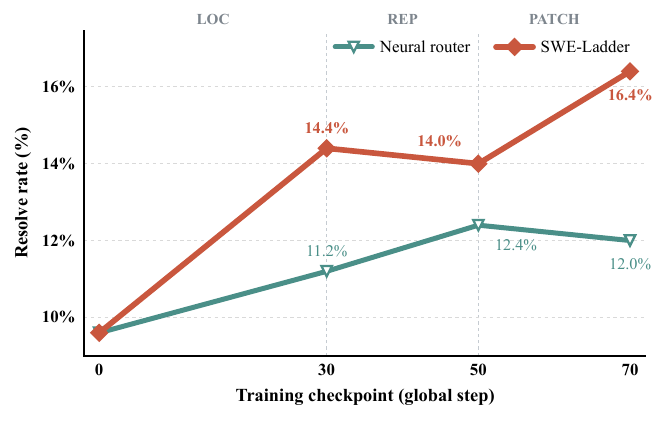}
  \caption{\textbf{Neural router versus \method evaluation performance.}
  Solve rates on SWE-bench Verified with the R2E-Gym scaffold. Both curves
  use the LOC$\rightarrow$REP$\rightarrow$PATCH training schedule and share
  a $9.6\%$ Qwen3-4B-Instruct-2507 base point; they differ only in the
  router, neural versus rule-based. Lines connect the
  reported evaluation points.}
  \label{fig:router-solve-comparison}
\end{figure}

Figure~\ref{fig:router-solve-comparison} reports the supplied neural-router
evaluation points alongside the complete rule-routed \method schedule. Despite
its lower label-agreement score, the rule-based router gives the stronger
downstream training result in this comparison, so annotation accuracy alone is
not a sufficient proxy for training utility.

\subsection{Evaluation Protocols}
\label{app:openhands-eval}

For the OpenHands rows in Table~\ref{tab:main-pspd}, we run the public
agent implementation and evaluate every configuration in the same
environment and protocol. The OpenHands scaffold and its evaluation harness
are kept fixed across the base, baseline, and \method rows, with a $128$K
context window, at most $150$ agent steps, and temperature $0.7$.
The R2E-Gym rows use the R2E-Gym scaffold and its evaluation harness, with a
$64$K context window, at most $150$ agent steps, and temperature $0.7$. The
two scaffold groups are reported separately in Table~\ref{tab:main-pspd}.

\FloatBarrier


\subsection{Termination-Aware Reward Shaping}
\label{app:reward-shaping}

For a fixed rollout, we suppress the indices $(u,g)$.
Let $V\in\{0,1\}$ be the verifier result, $\xi$ the termination
reason, and $e$ an indicator of a successful file-edit call,
used as a nonempty-patch proxy.
Let $\mathcal T_{\mathrm o}$ contain the soft-overlong reasons
\texttt{TRUNCATION}, \texttt{MAX\_STEPS}, and
\texttt{TRAJ\_TIMEOUT}. Let $\mathcal T_{\mathrm d}$ contain
\texttt{ENV\_DONE} and the empty reason $\varnothing$.
The reward shaping is
\begin{equation}
\begin{aligned}
R^{(0)}&=
\begin{cases}
0.5V, & \xi\in\mathcal T_{\mathrm o},\\
V-0.2, & \xi\in\mathcal T_{\mathrm d},\;V=e=0,\\
V, & \text{otherwise},
\end{cases}\\
R&=\operatorname{clip}(R^{(0)},-1,1).
\end{aligned}
\label{eq:implemented-reward-shaping}
\end{equation}
Thus, Eq.~\eqref{eq:implemented-reward-shaping} discounts a successful verifier
result from one to \(0.5\) under soft-overlong termination, while an unsolved
result remains zero.  The \(-0.2\) term applies
only when the agent voluntarily finishes (\texttt{ENV\_DONE} or an empty
reason), remains unsolved, and has made no successful edit.  Exploration-tool
reward and the no-tool-call penalty are both configured to zero in the main
training chain.

The overlong filter treats \texttt{NO\_CONTAINER},
\texttt{PROMPT\_TRUNCATION}, and \texttt{ENCODE\_ERROR} as hard failures.  For
these cases, the implementation clears the complete response mask instead of
treating the trajectory as an ordinary optimization sample.  After reward
shaping, group-relative advantages use the Bessel-corrected sample standard
deviation, with denominator \(G-1\), where \(G=4\) is the group size. GRPO is averaged over each rollout's
valid response tokens; OPSD is averaged over the subset assigned to the active
stage. A rollout with no active-stage response tokens contributes zero OPSD
loss.
The actor also includes the configured $0.005$ response-token entropy bonus in
the standard GRPO objective; this term is separate from privileged
distillation.

\subsection{Top-\texorpdfstring{$K$}{K} Distillation and Systems Configuration}
\label{app:topk-implementation}

Using the shorthand from
Section~\ref{sec:stage-conditioned-teacher}, let $p_t(v)$ and
$q_t(v)$ denote the student and teacher probabilities at the
same student-sampled prefix. For $v\in\mathcal K_t$, the
implementation computes
\begin{equation}
\log\widetilde q_t(v)
=\log q_t(v)
-\operatorname{LSE}_{w\in\mathcal K_t}\log q_t(w),
\end{equation}
where $\operatorname{LSE}$ denotes log-sum-exp over the indexed
tokens. The implemented token contribution is
\begin{equation}
\begin{aligned}
d_t(\theta)
&=\sum_{v\in\mathcal K_t}
\exp\!\bigl(\log p_t(v)\bigr)\\
&\quad{}\times\bigl[
\operatorname{sg}(\log p_t(v))
-\operatorname{sg}(\log\widetilde q_t(v))
\bigr].
\end{aligned}
\end{equation}
Both log terms are detached, so gradients flow through the
student probability factors in the surrogate. The teacher top-$K$
targets are aligned with the corresponding student token
probabilities at each sampled prefix.

The OPSD loss mask is the elementwise product of the valid model-response
mask and the active-stage indicator. The teacher's top-\(K\) targets are
conditioned on the active stage's privileged evidence and aligned to the
student-sampled response positions. Only selected positions contribute to
the loss, with no additional per-sample OPSD token cap. Targets at non-active
positions, if computed, are excluded from distillation. This stage-masked
objective is distinct from the historical full-token distillation variants.

The student vocabulary normalizer is accumulated in chunks, and probabilities
at the teacher token ids are gathered directly.  The reported eight-NPU setup
enables activation offload, FSDP parameter offload for the actor and frozen
reference policy, and optimizer offload for the actor.  Ulysses sequence
parallelism has degree eight and operates within each rollout sequence rather
than across rollouts.

GRPO and OPSD use per-rollout means over their respective response-token and
active-stage masks. The training hyperparameters are given in
Appendix~\ref{app:progressive-schedule}.

\subsection{Progressive Privilege Scheduling and Training Configuration}
\label{app:progressive-schedule}

\method applies privileged stages in the fixed order
$\mathrm{LOC}\rightarrow\mathrm{REP}\rightarrow\mathrm{PATCH}$.
Given phase budgets $U_L,U_R,U_P$, define the cumulative boundaries
$b_1=U_L$, $b_2=U_L+U_R$, and $b_3=U_L+U_R+U_P$.
The active training stage is
\begin{equation}
a(u)=
\begin{cases}
\mathrm{LOC}, & 0\leq u<b_1,\\
\mathrm{REP}, & b_1\leq u<b_2,\\
\mathrm{PATCH}, & b_2\leq u<b_3.
\end{cases}
\label{eq:progressive-schedule}
\end{equation}
Suppressing the trajectory index, the teacher context for
distillation is
\begin{equation}
c_{u,t}=c_{a(u)}.
\label{eq:scheduled-context}
\end{equation}
where $c_{a(u)}$ is the teacher context selected by the active stage, so
Eq.~\eqref{eq:scheduled-context} holds at every token of the phase. The
corresponding teacher predictions contribute to the loss only at tokens
selected by Eq.~\eqref{eq:stage-opsd-mask}. Thus, the LOC phase distills
LOC-labeled response tokens, the REP phase distills REP-labeled response
tokens, and the PATCH phase distills PATCH-labeled response tokens. No
unprivileged distillation term is applied to non-active stages. This schedule
changes the active-stage mask (and, when configured, the teacher context) while
preserving end-to-end rollouts and the full-response GRPO objective.

Each phase continuation warm-starts from its immediate predecessor, yielding a
single checkpoint sequence over the fixed schedule in
Eq.~\eqref{eq:progressive-schedule}.
The active privilege $a(u)$ is fixed for all rollouts collected at global
training iteration $u$.

\paragraph{Training configuration.}
We use $(U_L,U_R,U_P)=(30,20,20)$ global training iterations, $K=16$, and
$\lambda=0.15$. Each iteration samples 16 tasks with four rollouts per task,
yielding 64 rollouts, and the actor mini-batch size is one rollout.
The figures index completed training iterations from $1$: LOC occupies
steps $1$--$30$, REP steps $31$--$50$, and PATCH steps $51$--$70$.
This is equivalent to the zero-based update index $u$ in
Eq.~\eqref{eq:progressive-schedule}.

\paragraph{Inference.}
At inference, we deploy one end-to-end policy with the same rollout termination
conditions: submission, environment completion, or the configured resource
limit. Inference uses no teacher, no stage-specific deployed policies, and no
extra trainable parameters.

\section{Stage-Routed Privileged Information}
\label{app:prompts}

\begin{figure*}[!t]
\centering
\includegraphics[width=0.9\linewidth]{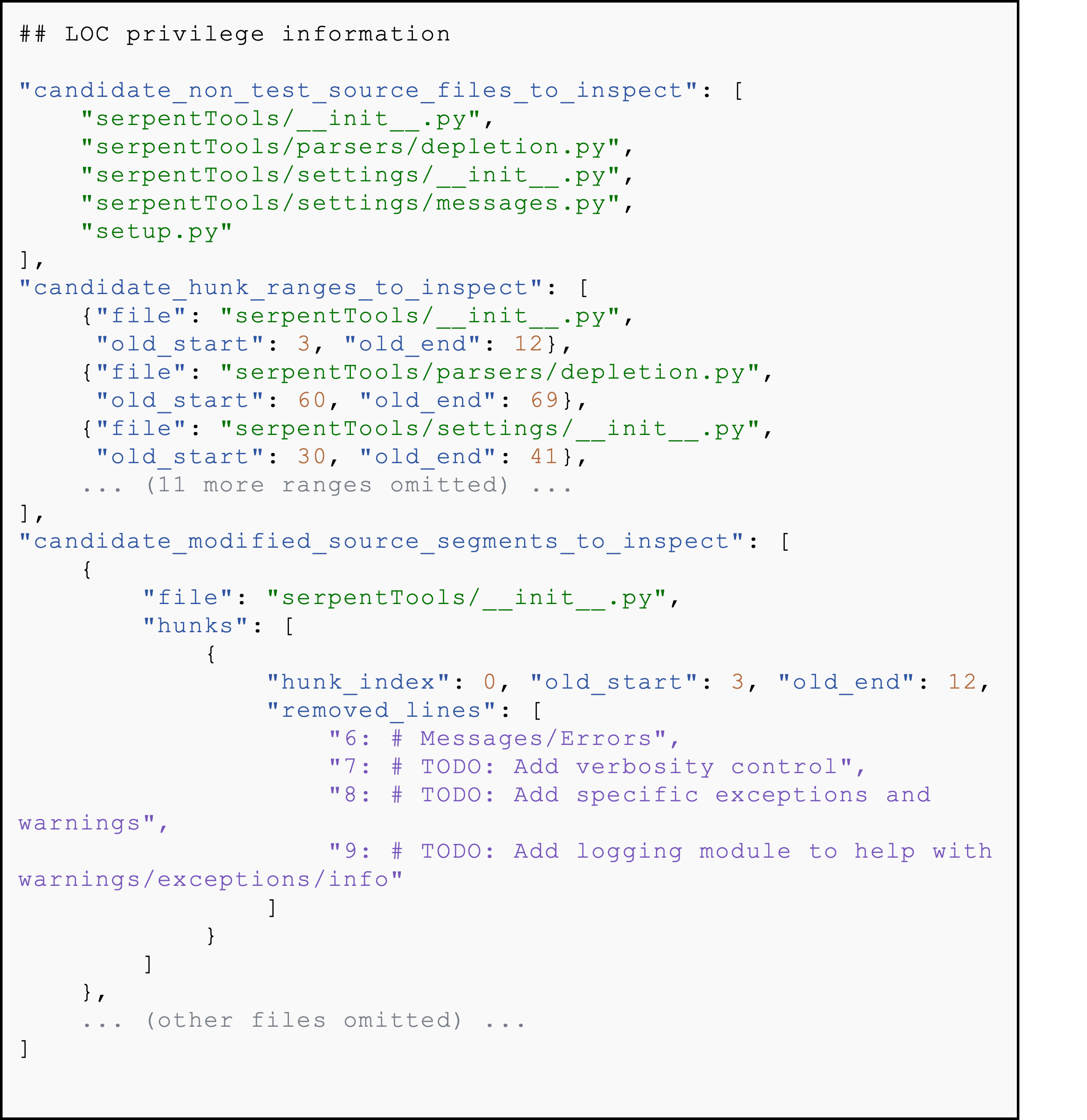}
\caption{\textbf{Localization privileged information.} Candidate files,
old-side hunk ranges, and source fragments identify where the issue should be
investigated without revealing the replacement patch.}
\label{fig:pi-loc}
\end{figure*}

\begin{figure*}[!t]
\centering
\includegraphics[width=0.9\linewidth]{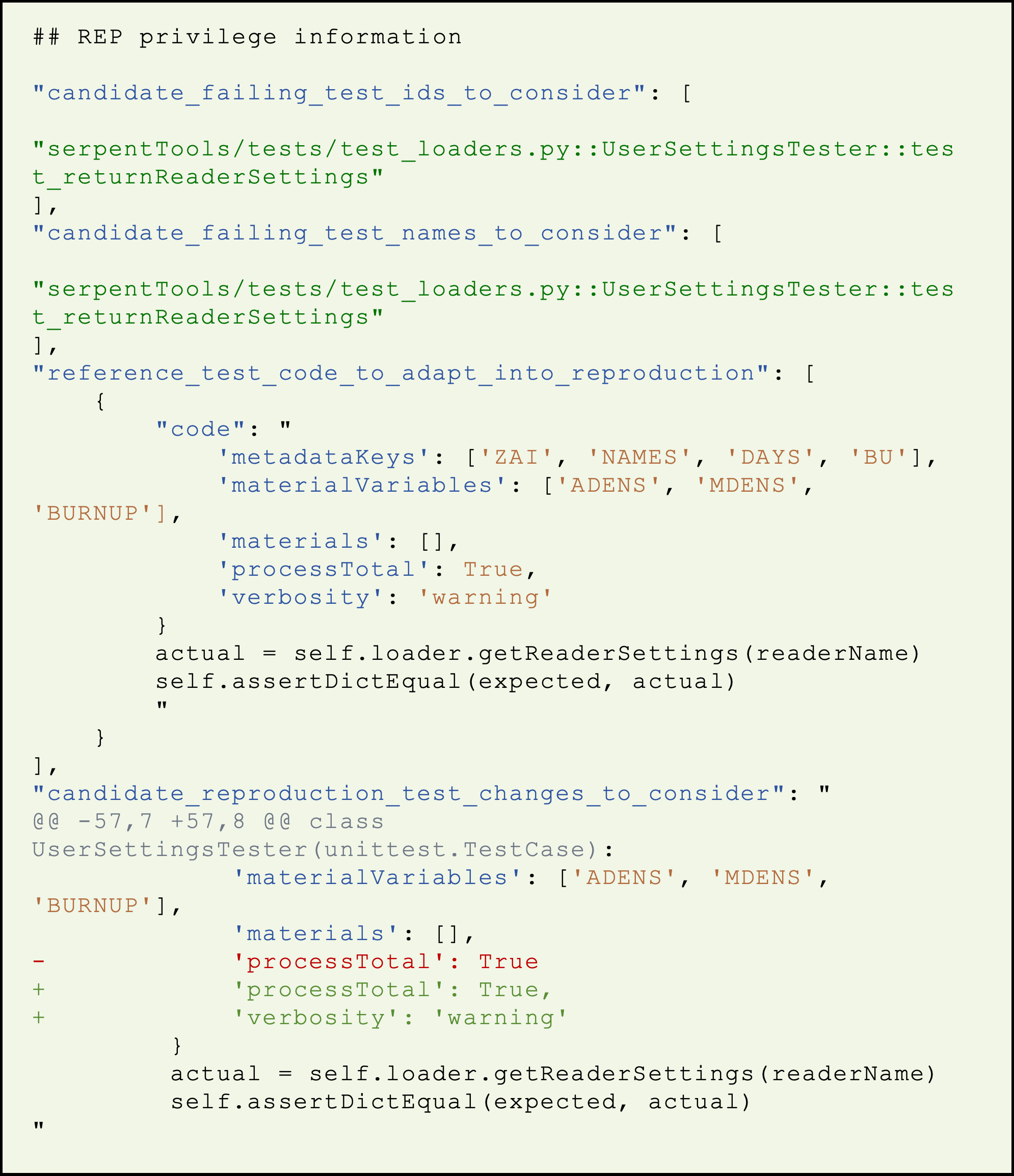}
\caption{\textbf{Reproduction privileged information.} Fail-to-pass tests,
reference test snippets, and candidate test changes describe the failing
behavior and provide a starting point for reproducing it.}
\label{fig:pi-repro}
\end{figure*}

\begin{figure*}[!t]
\centering
\includegraphics[width=0.9\linewidth]{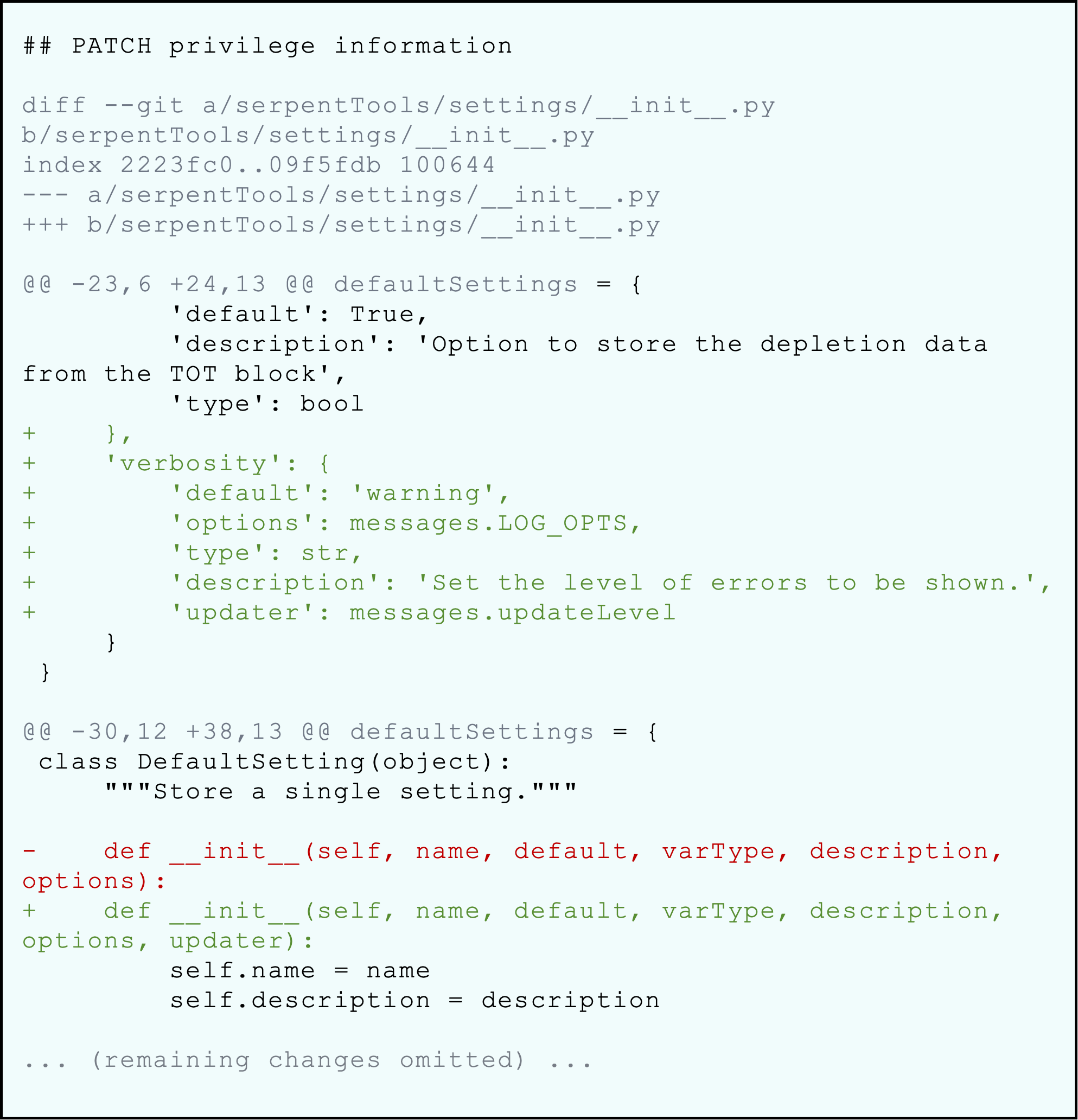}
\caption{\textbf{Patch privileged information.} The verified repair diff
specifies the intended source-code correction available to the patch-stage
teacher.}
\label{fig:pi-patch}
\end{figure*}

The main text describes how the three stage-specific evidence views are
constructed. This appendix gives a representative router trace
(Table~\ref{tab:router-trace}), the evidence fields supplied for each stage
(Table~\ref{tab:privileged-examples} and
Figures~\ref{fig:pi-loc}--\ref{fig:pi-patch}), and the token cost that each
evidence block adds to the teacher prompt
(Table~\ref{tab:privilege-cost}).

\begin{table}[!ht]
\centering
\small
\caption{Schematic examples of the stage-specific privileged information.
The actual fields are populated from each instance's verified repair metadata.}
\label{tab:privileged-examples}
\begin{tabular}{@{}p{0.18\columnwidth}p{0.74\columnwidth}@{}}
\toprule
Stage & Example privileged fields \\
\midrule
\textsc{LOC} &
\texttt{changed file: src/module.py}; old-side hunk range; corresponding
pre-patch source lines. \\
\textsc{REP} &
Fail-to-pass test identifier and name; reference test code; candidate test
change or reproduction starting point. \\
\textsc{PATCH} &
Verified repair diff for the changed source file, including the intended
replacement code. \\
\bottomrule
\end{tabular}
\end{table}

\begin{table*}[b]
\centering
\small
\caption{A representative $21$-step trajectory (an \texttt{oauthlib} issue),
labeled by the rule-based router of Appendix~\ref{app:stage-annotation}.
Re-running the reproduction script while in \textsc{PATCH} demotes the state
to \textsc{REP} (step 12), and the third consecutive inspection step returns
it to \textsc{LOC} (step 15), so the label sequence is not monotone.}
\label{tab:router-trace}
\scriptsize
\setlength{\tabcolsep}{3pt}
\begin{tabular}{@{}r p{0.14\textwidth} p{0.22\textwidth} p{0.34\textwidth} p{0.16\textwidth}@{}}
\toprule
Step & Tool & Argument & Path & Assigned stage \\
\midrule
1--3 & \texttt{search} & \path{nonce, add_id_token}, \ldots & \path{/testbed} & localization \\
4--7 & \texttt{file\_editor} & \texttt{view} & \path{openid/connect/core/grant_types/} & localization \\
8 & \texttt{file\_editor} & \texttt{create} & \path{/testbed/reproduce_issue.py} & reproduction \\
9 & \texttt{execute\_bash} & \path{python3 reproduce_issue.py} & --- & reproduction \\
10--11 & \texttt{file\_editor} & \texttt{str\_replace} & \path{openid/connect/core/grant_types/} & patch generation \\
12 & \texttt{execute\_bash} & \path{python3 reproduce_issue.py} & --- & reproduction \\
13--14 & \texttt{file\_editor} & \texttt{view} & \path{openid/connect/core/request_validator.py} & reproduction \\
15 & \texttt{file\_editor} & \texttt{view} & \path{openid/connect/core/tokens.py} & localization \\
16--20 & \texttt{file\_editor} / \texttt{execute\_bash} & \texttt{str\_replace} / run & \path{/testbed/reproduce_issue.py} & reproduction \\
21 & \texttt{finish} & \texttt{submit} & --- & reproduction \\
\bottomrule
\end{tabular}
\end{table*}

\begin{table*}[b]
\centering
\small
\caption{Tokens added by each privileged evidence block over the shared task
prompt across the $1{,}149$ training instances (Qwen3-4B-Instruct-2507 tokenizer).}
\label{tab:privilege-cost}
\begin{tabular}{lrrrr}
\toprule
Evidence block & median & mean & p90 & max \\
\midrule
\texttt{causal\_localization} & $406$ & $544$ & $987$ & $8{,}836$ \\
\texttt{localized\_issue\_reproduction} & $1{,}249$ & $2{,}928$ & $7{,}084$ & $32{,}604$ \\
\texttt{patch\_generation} & $669$ & $907$ & $1{,}864$ & $9{,}893$ \\
\midrule
union of all three (unrouted variant) & $2{,}709$ & $4{,}375$ & $9{,}592$ & $36{,}173$ \\
\bottomrule
\end{tabular}
\end{table*}

\subsection{Representative Privileged-Information Views}

The following examples show the evidence blocks supplied to the frozen
teacher for the three repair stages. They are rendered from the structured
fields described in the main text; the student prompt does not include these
blocks.

\section{Additional Training Diagnostics}
\label{app:training-curves}

\subsection{Training Dynamics}
\label{app:dynamics-pspd}


Figure~\ref{fig:dynamics-audited} compares GRPO, GRPO+OPSD,
SDAR, and the complete \method sequence over a common $70$-iteration horizon.
We distinguish model-generated tokens from tool/environment output and plot
shaped training reward separately from held-out pass@1. Curves show individual
training runs, not averages over independent seeds.

The GRPO run exhibits a pronounced decline in its optimization diagnostics.
Comparing iterations $1$--$10$ with $61$--$70$, its mean entropy diagnostic
falls from $0.0300$ to $0.0061$, mean gradient norm from $1.820$ to $0.187$,
and mean training reward from $0.236$ to $0.055$. Meanwhile, mean generated
tokens increase from $8.29$k to $16.73$k. \method's corresponding entropy means
are $0.0251$ and $0.0221$, and its gradient-norm means are $2.353$ and
$2.397$. Its mean generated length increases from $10.09$k to $11.62$k.
This sequence does not exhibit the same diagnostic decline as the GRPO run,
although its reward remains noisy and does not increase monotonically.

The other privileged baselines provide an important qualification. The
GRPO+OPSD and SDAR traces retain additional optimization signal
late in training, but preserving entropy or gradient magnitude alone does not
explain \method's held-out performance. These plots therefore do not establish
a general failure of either baseline; they provide diagnostic context for the
stage-aligned comparison.
Appendix~\ref{app:curve-reconstruction} defines the plotted metrics.

\subsection{Metric Definitions}
\label{app:curve-reconstruction}

The panels report logged actor entropy and gradient norm, the mean number of
model-generated tokens, and the mean shaped training reward; the plotting
manifest records their exact source fields. We retain the implementation's
logged scale for entropy
and gradient norm rather than treating them as independently recalculated
policy statistics. In particular, generated-token counts exclude the
environment output included in the broader \texttt{response\_length/mean}
field. The reward includes termination-aware shaping
(Appendix~\ref{app:reward-shaping}) and must not be read as
SWE-bench pass@1. The teacher-top-$K$ loss is not plotted as a calibrated KL
distance for the reasons discussed in Appendix~\ref{app:topk-implementation}.

\end{document}